\documentclass[11pt, a4paper]{article}
\usepackage[hyperref]{eacl2021}
\usepackage{latexsym}

\usepackage{graphicx}
\usepackage{tipa}    %for ipa
\usepackage{pifont}    %for pointing hand
\usepackage[T1]{fontenc}
\usepackage[utf8]{inputenc}
\usepackage[russian, ethiop, main=english]{babel}
\usepackage{cfr-lm}	%for old style numerals
\usepackage{amssymb}
\usepackage{enumerate}
\usepackage{times} %for Times New Roman
\usepackage{amsmath}
\usepackage[normalem]{ulem}
\usepackage{tabularx}
\usepackage{tikz}
\usepackage{tikz-qtree}
\usepackage{mdwlist}
\usepackage{paralist}
\usepackage{skull}
\usepackage{multicol}
\usepackage{multirow}
\usepackage{upgreek}
\usepackage{color}
\usepackage{rotating}	%for sideways tables
\usepackage{enumerate}
\usepackage{tabu}
\usepackage{booktabs}
\usepackage{linguex}
\usepackage{natbib}
\usepackage{hyperref}
\hypersetup{colorlinks=true,citecolor=blue, linkcolor=blue, urlcolor=blue}
\usepackage{hyphenat}
\usepackage{datetime}
\usepackage{dirtree}

\usepackage{microtype}

\aclfinalcopy % Uncomment this line for the final submission
\def\aclpaperid{23} %  Enter the acl Paper ID here

\title{Text Normalization for Low-Resource Languages of Africa}

\author{Andrew Zupon\footnotemark[1]\\
  University of Arizona\\
  \texttt{zupon@email.arizona.edu} \\\And
  Evan Crew\\
  Google\\
  \texttt{crewe@google.com}\\\And
  Sandy Ritchie\\
  Google\\
  \texttt{sandyritchie@google.com}}

\date{}

\begin{document}
\renewcommand{\thefootnote}{\fnsymbol{footnote}}
\maketitle
\begin{abstract}
Training data for machine learning models can come from many different sources, which can be of dubious quality.
For resource-rich languages like English, there is a lot of data available, so we can afford to throw out the dubious data.
For low-resource languages where there is much less data available, we can’t necessarily afford to throw out the dubious data, in case we end up with a training set which is too small to train a model.
In this study, we examine the effects of text normalization and data set quality for a set of low-resource languages of Africa---Afrikaans, Amharic, Hausa, Igbo, Malagasy, Somali, Swahili, and Zulu.
We describe our text normalizer which we built in the Pynini framework, a Python library for finite state transducers, and our experiments in training language models for African languages using the Natural Language Toolkit (NLTK), an open-source Python library for NLP.
\end{abstract}

\footnotetext[1]{Work performed as an intern at Google.}

\renewcommand{\thefootnote}{\arabic{footnote}}

\section{Introduction}
\label{sec:intro}

The goals of this project are to study the effects of text normalization and the quality of existing open source data sets for a set of low-resource languages of Africa.
This involves building a text normalizer using Pynini \citep{gorman-2016-pynini}, an open-source Python library for finite-state grammars, and training a language model using the Natural Language Toolkit (NLTK), an open-source Python library for NLP.
To reach these goals, we perform experiments with this text normalizer to answer two main questions:

\begin{enumerate}[Q1.]
	\item How does the inclusion of language-specific rewrite rules to the text normalizer affect the perplexity of the language model?
	
	\item Can we evaluate the quality of a data set by using the proportion of sentences rejected by the text normalizer as a proxy?
\end{enumerate}

The impact of this project is multifold.
The languages chosen for this project have over 300 million native speakers, many of whom are coming online now and in the next few years.
Their ability to communicate and access information online may be limited if their native languages are not supported in the applications which they use to access this information.
This presents a double edged sword for both developers and users; the lack of representation online for these languages and many others means that there is a lack of natural language data available.
Much of what exists online for these low-resource languages is not representative of everyday language use either: instead it primarily consists of news and Wikipedia articles, religious texts, etc.
Additionally, many low-resource languages have conflicting orthographic standards from historical script changes, to new national standards.
By spending some time to understand the specific linguistic and historical situations of these languages, we can more effectively develop technology to support them.

\section{Text Normalization}

Text normalization is the process of transforming text into a single canonical form.
Normalization can be used to combine words with variable diacritics (<r\'esum\'e>\footnote{Graphemes and text strings are represented here between <angled brackets>.} vs <resume>), capitalization (<John> vs <john>), or spelling convention (U.S.\ <color> vs U.K.\ <colour>) into a single form.

In addition, normalization can be used to separate or remove punctuation marks that may not be relevant to the task.
For example, for training a simple part of speech tagger, we would likely want to not distinguish between the sequence <John arrived.> and <John arrived>, even though the former has a <.> attached to the second token.

This is also useful for applications including text-to-speech, where non-word text can be pronounced differently depending on the language or context.
For example, the text string <\$200> can be pronounced as ``two hundred dollars'' in English but ``deux cents dollars'' in French.
Normalizing this text to the appropriate canonical form based on the language can help the system yield the correct pronunciation when it encounters both <\$200> and <two hundred dollars>.

\section{Related Work}

There is previous work on text normalization in general and some work on text normalization for African languages in particular.

The text normalizer in this work is based loosely on \cite{46952}, which implements a text normalization system in Thrax, a different open-source library for finite-state grammars.

In addition to general work on text normalization, there has been work on African languages in particular.
This work includes work on individual languages, such as Yoruba \cite{asahiah2017restoring}, Swahili \cite{hurskainen2004swahili}, and Zulu \cite{pretorius2003computational}, as well as some crosslinguistic \cite{peche2009development}.
However, much of the crosslinguistic work focuses on languages of South Africa, which ignores many of the languages we are interested in.

\section{Languages}

This project considers eight low-resource languages of Africa.
The languages come from different language families and different geographical regions.
Table~\ref{table:languages} provides information about these languages, including language family, number of speakers, and the countries they are primarily spoken in.

\begin{table*}[tb]
	\small
	\begin{center}
		\begin{tabu} to \linewidth {X[1,l]X[2.5,l]X[1,l]X[1.5,l]}\toprule
			\multicolumn{1}{c}{\textbf{Language}} & \multicolumn{1}{c}{\textbf{Family}} & \multicolumn{1}{c}{\textbf{Speakers}} & \multicolumn{1}{c}{\textbf{Location}}\\\midrule
			\rule{0pt}{2ex}Afrikaans	&	Indo-European, Germanic	&	15--23 M	&	South Africa, Namibia\\%\midrule
			%			&&&\\
			\rule{0pt}{3.5ex}Amharic	&	Afro-Asiatic, Semitic, South Semitic	&	22 M	&	Ethiopia\\%\midrule
			%			&&&\\
			%			\rule{0pt}{3.5ex}Bambara	&	Niger-Congo, Mande	&	15 M	&	Mali\\%\midrule
			%			&&&\\
			\rule{0pt}{3.5ex}Hausa	&	Afro-Asiatic, Chadic	&	100--150 M	&	Niger, Nigeria\\%\midrule
			%			&&&\\
			\rule{0pt}{3.5ex}Igbo	&	Niger-Congo, Atlantic-Congo, Volta-Congo, Volta-Niger, Igboid	&	35 M	&	Nigeria\\%\midrule
			%			&&&\\
			\rule{0pt}{3.5ex}Malagasy	&	Austronesian, Malayo-Polynesian	&	25 M	&	Madagascar\\%\midrule
			%			&&&\\
			\rule{0pt}{3.5ex}Somali	&	Afro-Asiatic, Cushitic	&	16 M	&	Somalia, Somaliland, Djibouti, Ethiopia, Kenya\\%\midrule
			%			&&&\\
			\rule{0pt}{3.5ex}Swahili	&	Niger-Congo, Atlantic-Congo, Volta-Congo, Bantu, Northeast Bantu	&	100--150 M	&	Tanzania, Uganda, Rwanda, Burundi, Kenya\\%\midrule
			%			&&&\\
			%			\rule{0pt}{3.5ex}Wolof	&	Niger-Congo, Atlantic-Congo, Atlantic	&	10 M	&	Senegal, Gambia, Mauritania\\%\midrule
			%			&&&\\
			%			\rule{0pt}{3.5ex}Yoruba	&	Niger-Congo, Atlantic-Congo, Volta-Congo, Volta-Niger, Yoruboid	&	45-50 M	&	Nigeria, Benin, Togo\\%\midrule
			%			&&&\\
			\rule{0pt}{3.5ex}Zulu	&	Niger-Congo, Atlantic-Congo, Volta-Congo, Bantu, Southern Bantu\vspace{4pt}	&	28 M	&	South Africa, Lesotho, Eswatini\\\bottomrule
		\end{tabu}
	\end{center}
	\caption{Languages chosen for this project, including their (approximate) language family, approximate total number of speakers, and the primary countries they are spoken in.}
	\label{table:languages}
\end{table*}

\section{Data}

For any NLP task, the source of the data plays an important role.
One could use existing corpora or create a new data set specific to the project.
Since one of the goals of this project is to evaluate the quality of existing data sets, we do not create any new corpora.\footnote{Though our results on some languages suggest that would be a good next step.}

For this project, we looked for data sets that would (1) benefit from normalization and (2) likely be useful for training a language model.
For these reasons, we excluded corpora that were too clean/manicured---not needing normalization---and corpora that were simply lists of words---not useful for a language model.

The sources we chose are Universal Dependencies (UD)\footnote{\href{https://universaldependencies.org/}{https://universaldependencies.org/}}, the Leipzig Corpora Collection (LCC)\footnote{\href{https://wortschatz.uni-leipzig.de/en/}{https://wortschatz.uni-leipzig.de/en/}}, the OSCAR corpus (OSCAR)\footnote{\href{OSCAR-corpus.com}{OSCAR-corpus.com}}, and the An Cr\'ubad\'an corpus (AC)\footnote{\href{crubadan.org}{crubadan.org}}.
We initially also considered UniMorph\footnote{\url{https://unimorph.github.io/}}, but due to the data being exceptionally clean and consisting of single words/lemmas, we chose to exclude it from our experiments.

Table~\ref{table:corpora} summarizes the availability of data for each language considered from each source.
With the exception of An Cr\'ubad\'an, no source covers all languages. Only Afrikaans and Amharic are covered by all sources we chose to include.

\begin{table}[tb]
	%\small
	\begin{center}
		\begin{tabu} to \linewidth {lcccc} \toprule
			\multirow{2}{*}{\textbf{Language}} & \multicolumn{4}{c}{\textbf{Data Source}}\\\cline{2-5}
			{} & UD & LCC & OSCAR & AC\\\midrule
			Afrikaans	&	\ding{51}	&	\ding{51}	&	\ding{51}	&	\ding{51}\\
			Amharic		&	\ding{51}	&	\ding{51}	&	\ding{51}	&	\ding{51}\\
			%			Bambara		&	\ding{51}	&		&		&	\ding{51}\\
			Hausa		&		&		&		&	\ding{51}\\
			Igbo		&		&		&		&	\ding{51}\\
			Malagasy	&		&	\ding{51}	&	\ding{51}	&	\ding{51}\\
			Somali		&		&	\ding{51}	&	\ding{51}	&	\ding{51}\\
			Swahili		&		&	\ding{51}	&	\ding{51}	&	\ding{51}\\
			%			Wolof		&	\ding{51}	&		&		&	\ding{51}\\
			%			Yoruba		&	\ding{51}	&	\ding{51}	&	\ding{51}	&	\ding{51}\\
			Zulu		&		&	\ding{51}	&		&	\ding{51}\\\bottomrule
		\end{tabu}
	\end{center}
	\caption{
		Data available from Universal Dependencies (UD), Leipzig Corpora Collection (LCC), OSCAR (OSCAR), and An Cr\'ubad\'an (AC).
	}
	\label{table:corpora}
\end{table}

The Universal Dependencies\footnote{Version 2.6, released 02020.05.15.} data consists of sentences from various sources annotated with morphosyntactic information including lemma, part of speech, and dependency relations.

Because the Universal Dependencies data is (usually) annotated by hand, these data sets are often much smaller than data sets from the other sources (which are mostly unannotated).

The Leipzig Corpora Collection is a collection of corpora from various sources on the web, including Wikipedia, newswire, and other web sources.
The downloadable corpora are often available in different sizes and come from different sources.
%Table~\ref{table:LCC} shows the largest corpus available to download for each language from each genre.
Only Afrikaans has a data set from each genre, and Swahili is split into two categories:\ \ Swahili (individual language) and Swahili (macrolanguage). %\footnote{The numbers for Swahili in Table~\ref{table:LCC} are inclusive of both sets of corpora.}.
The size of available corpora also varies greatly from 10K sentences (Swahili Newscrawl) to 1M sentences (Afrikaans Web and Mixed).

The OSCAR\footnote{OSCAR is constantly expanding. The data used for this project was downloaded on 2020.09.09.} corpus is a multilingual corpus of data obtained by language classification and filtering of the Common Crawl corpus.
The downloadable data is available in ``original'' and ``deduplicated'' versions.

An Cr\'ubad\'an is crawled from various sources on the web, including Wikipedia, Bible translations, blogs, and Twitter.
The downloadable data consists of two files:\ \ one file is a list of words in the language sorted by frequency, the second file is a list of word bigrams in the language sorted by frequency.

\section{Project Pipeline}

The pipeline of this project consists of two primary components:\ \ the text normalizer and the language model.\footnote{The code for this project can be found at \url{https://github.com/googleinterns/text-norm-for-low-resource-languages}.}

\subsection{Text Normalizer}

The text normalizer is written using Pynini, an open-source Python library for finite-state grammars \cite{gorman-2016-pynini}.
The text normalizer applies the following six steps from \cite{46952}:
\vfill%\null
%\vspace{-8pt}

\begin{enumerate}
	\itemsep=-4pt
	\item {Language-agnostic preprocessing\footnote{
			This step lowercases text, applies Unicode NFC normalization, and applies language-universal rules (e.g. replacing all apostrophe-like characters with APOSTROPHE).
			
			Unicode NFC (Normalization Form C) is Canonical Decomposition followed by Canonical Composition.
			This will take a grapheme like <\'o>, decompose it into <o>+<{ \'{}}>, and then recombine it into <\'o>.
			This is useful because some graphemes are one Unicode character and other identical-looking graphemes are multiple composed Unicode characters, so <\'o> $\neq$ <\'o> every time.
	}}

	\item {Invalid token filtering}\footnote{
		This step rejects entire sentences that contain tokens not valid for the language.
		Valid tokens consist of optional initial punctuation, followed by alphanumeric characters, followed by final punctuation, as well as some limited times, large numbers, and web/email addresses.
		This approach is subjective, and will undoubtedly miss some valid tokens, but we want to be conservative in what we allow.}

	%	\vspace{-6pt}
	\item {Applying language-specific rules}
	
	%	\vspace{-6pt}
	\item {Detaching punctuation\footnote{Notably we keep APOSTROPHE (\texttt{\textquotesingle}) and HYPHEN ({-}), which can be counted as graphemes in some languages.}}
	
	%	\vspace{-6pt}
	\item Deleting freestanding punctuation
	
	%	\vspace{-6pt}
	\item {Deleting extra whitespace\footnote{More than one single space between tokens}}
\end{enumerate}

Table~\ref{table:example} shows an example derivation demonstrating how the text normalizer applies these six steps to a sample sentence of Malagasy with a single Russian word thrown in.
Note that in Malagasy, <@> is a common abbreviation for <amin'ny> `with the'.

\begin{table*}[tb]
	% \small
	\begin{center}
		\begin{tabu} to \linewidth {cll}\toprule
			\multicolumn{2}{c}{\textbf{Normalization Step}} & \multicolumn{1}{c}{\textbf{Text}}\\\midrule
			{} & INPUT & \texttt{<}\foreignlanguage{russian}{Собака} @ FIRY IZAO?\texttt{>}\\
			1 & Language-agnostic preprocessing: & \texttt{<}\foreignlanguage{russian}{собака} @ firy izao?\texttt{>}\\
			2 & Token/sentence filtering & \texttt{<}<UNK> @ firy izao?\texttt{>}\\
			3 & Applying language-specific rules & \texttt{<}<UNK> amin'ny firy izao?\texttt{>}\\
			4 & Detaching punctuation & \texttt{<}<UNK> amin'ny firy izao ?\texttt{>}\\
			5 & Deleting freestanding punctuation & \texttt{<}<UNK> amin'ny firy izao \texttt{>}\\
			6 & Deleting extra whitespace & \texttt{<}<UNK> amin'ny firy izao\texttt{>}\\
			{} & OUTPUT & \texttt{<}<UNK> amin'ny firy izao\texttt{>}\\\bottomrule
			
		\end{tabu}
		
		\caption{Sample application of text normalizer to a sentence of Malagasy.
			The initial and final \texttt{<} and \texttt{>} indicate the boundaries of the text and are not actually part of the string. These illustrate the extra whitespace at the end of the string in step 5.}
		\label{table:example}
	\end{center}
\vspace{-12pt}
\end{table*}

\subsection{Language Model}

The language model is written using the Natural Language Toolkit (NLTK), an open-source Python library for natural-language processing \cite{bird2009natural}.
The language model reads in the output of the normalizer, partitions the data into training and testing partitions (an 80-20 split), fits a language model to the data, and calculates the average perplexity across ngrams.

Perplexity (\textit{PP}) is the inverse of the probability of a test set, normalized by the number of words.
For a test set \textit{W} with a bigram language model:
\begin{equation}
PP(W) = \sqrt[N]{\prod_{i=1}^{N}\frac{1}{P(w_{i}|w_{1})}}
\end{equation}

The lower the perplexity, the better the probability distribution is at predicting the sample.
We chose language modeling as the task and perplexity as the evaluation metric here based on the data available.
Other tasks such as parsing or text-to-speech require labeled text or audio data.
Language modeling does not require any labeled data, which makes it possible given the data we have to work with for these languages.
Even so, perplexity has its drawbacks.
One way to guarantee lower perplexity would be to replace all consonants and vowels with <C> and <V>, respectively, or replace every word with <uh>.
This would necessarily reduce the perplexity, but would lose important and useful distinctions in the language.
While our experiments use perplexity as one evaluation metric, a decrease in perplexity is not necessarily an improvement in the model.
Rather, it provides a cue as to if our language-specific rules are being used, which can inform us which data sets to include later in the pipeline.

Our language model uses bigrams and unigrams.
With sentence-based filtering---Step 2 of the text normalizer, where whole sentences containing invalid tokens are replaced by a placeholder token---the rejected sentences are filtered out before partitioning the data.
To deal with unseen tokens and ngrams, the language model uses Laplace (or add-one) smoothing \cite{manning2008introduction}.
Other smoothing approaches were not considered.
In the end, we calculate the average perplexity across ngrams in the testing partition.

\section{Experiments}

As discussed in Section\ \ref{sec:intro}, our experiments are designed to answer two questions:
\begin{enumerate}[Q1.]
	\item How does the inclusion of language-specific rewrite rules to the text normalizer affect the perplexity of the language model?
	\item Can we evaluate the quality of a data set by using the proportion of sentences rejected by the text normalizer as a proxy?
\end{enumerate}

To answer Q1, we compare a \texttt{base} normalizer---one that includes only the language-agnostic steps---with an \texttt{experiment} normalizer---one that adds the language-specific rules.

If the perplexity decreases between the base and experiment normalizers, then the language-specific rules had an effect.
Based on preliminary results, perplexity appears to be roughly linear with respect to the number of ngrams, these experiments also consider the relative decrease in perplexity.
This allows us to compare experiments across languages and data sets, even when the size of those data sets varies.\footnote{For example, perplexity the LCC 30K data sets is roughly twice as high as for the LCC 10K data sets. We want to be able to compare these even with this raw difference.}
We can then compare this relative change with our other experiments to see which language-specific rules or data sets yielded a greater contribution.

To answer Q2, we consider the number of sentences accepted and rejected by the normalizer.
That is, we count how many sentences Step 2 of the sentence-based text normalizer rejects.
The higher the proportion of rejected sentences, the noisier the data set is likely to be.
One caveat is that this proportion is a reflection of how we define valid/invalid sequences.
For example, we chose to cap the length of digit sequences to disallow extremely long numbers.\footnote{
	In our case, we allow up to six digits (with digit separators) followed by up to four decimal places.}
This is not a perfect solution, but it allows to limit the space of valid sequences for experimental purposes.
Given the proportion of rejected sentences, we can evaluate which languages and corpora have data that might be more useful for a future task.

The following sections discuss the language-specific experiments we ran, followed by a discussion of results for the language-specific experiments and the quality of the data sets.

\subsection{Amharic Overdifferentiation}

Amharic is the only language of our set that does not primarily use a Latin-based orthography, instead the Ge`ez abugida is used \cite{comrie2009world}.
Amharic inherited this orthography from the Ge`ez language---a closely-related South Semitic language still used as a liturgical language.

However, due to historical sound changes, some of the graphemes that represent different sounds in Ge`ez are now pronounced the same in Amharic.
For example, the phoneme /h/ can be represented by the <\foreignlanguage{ethiop}{ha}>, <\foreignlanguage{ethiop}{.ha}>, <\foreignlanguage{ethiop}{_ha}> and <\foreignlanguage{ethiop}{'ha}> series of graphemes; the phoneme /ts/ can be represented by the <\foreignlanguage{ethiop}{.ca}> and <\foreignlanguage{ethiop}{.sa}> series of graphemes; and the phoneme /\textipa{P}/ can be represented by the <\foreignlanguage{ethiop}{'a}> and <\foreignlanguage{ethiop}{`a}> series of graphemes.

\citet{menuta2016over} surveyed Amharic speakers and discovered that people vary both in their preference of a particular grapheme series but also in their actual use.
\citeauthor{menuta2016over} points out that for the /ts/ case, survey respondents strongly preferred the <\foreignlanguage{ethiop}{.ca}> series---94.1\% compared to 5.9\% for the <\foreignlanguage{ethiop}{.sa}> series---but a textual analysis shows that people use the <\foreignlanguage{ethiop}{.sa}> series more often than the <\foreignlanguage{ethiop}{.ca}> series.

For the present experiment, we reduce the overdifferentiation of Amharic graphemes by collapsing the different grapheme series for /h/, /ts/, and /\textipa{P}/ to one series each.
This experiment chooses the most preferred variant based on \citeauthor{menuta2016over}'s surveys.
The different grapheme series for each of these phonemes and the preferred variant are given in Table~\ref{table:amharic}. %---in this case, <\foreignlanguage{ethiop}{ha}> for /h/, <\foreignlanguage{ethiop}{.c}> for /ts/, and <\foreignlanguage{ethiop}{'a}> for /\textipa{P}/.
Future work could consider instead the more commonly used variant based on textual analysis.\footnote{\citep{menuta2016over} mentions textual analysis for the case of [ts], but does not provide sufficient details overall to use it here.}

\begin{table}[tb!]
	\begin{center}
		
		\begin{tabu} to \linewidth {ccccc}\toprule
			\multirow{2}{*}{\textbf{Phoneme}} & \textbf{Preferred} & \multicolumn{3}{c}{\textbf{Other}}\\
			{} & \textbf{Series} & \multicolumn{3}{c}{\textbf{Series}}\\\midrule
			/h/ & <\foreignlanguage{ethiop}{ha}> & <\foreignlanguage{ethiop}{.ha}> & <\foreignlanguage{ethiop}{_ha}> & <\foreignlanguage{ethiop}{'ha}>\\
			/ts/ & <\foreignlanguage{ethiop}{.ca}> & & <\foreignlanguage{ethiop}{.sa}> &\\
			/\textipa{P}/ & <\foreignlanguage{ethiop}{'a}> & & <\foreignlanguage{ethiop}{`a}> &\\\bottomrule
		\end{tabu}
		
	\end{center}
	\caption{Amharic phonemes with the preferred grapheme series and other grapheme series, based on surveys by \citet{menuta2016over}.}
	\label{table:amharic}
\end{table}

\subsection{Zulu Hyphenation}

As a Bantu language, Zulu has a rich system of noun classes and accompanying noun classifiers.
Noun classifiers are typically written attached to the following word, as in <isiZulu> `Zulu language'.
In contrast, with vowel-initial loanwords, a hyphen is sometimes inserted between the classifier and the noun, as in <i-Afrika> `Africa'.
However, the use of a hyphen here is inconsistent.
For example, the Zulu language Wikipedia page for Africa\footnote{\url{https://zu.wikipedia.org/wiki/IAfrika}, as of 02020.09.10} is titled <iAfrika>, without a hyphen, but includes <i-Afrika> in the body of the text.
Our experiment looks to unify these forms by removing hyphens between noun classifiers and vowel-initial words.

\subsection{Malagasy Substitution \& Abbreviation}

Our Malagasy experiments evaluate the normalization of two distinct phenomena.
First, the official Malagasy orthography includes the grapheme <\"n> \cite{biddulph1997introduction}.
Due to the rarity of this grapheme\footnote{The grapheme <\"n> is only found in about seven languages and the name of the band Sp\i\"nal Tap.}, it is not frequently available on computer or mobile phone keyboards.
Instead, Malagasy speakers will substitute <\~n>, which is much more commonly available.
Second, Malagasy speakers often use the abbreviation <@> for the common word <amin'ny> `with the'.

For our experiments, we replace all instances of <\~n> with <\"n> and replace all instances of standalone <@>---that is, not part of email addresses and such---with the full form <amin'ny>.

\subsection{Afrikaans Contractions}

Afrikaans has three apostrophe-initial words <'n>, <'t>, and <'k>.
<'n> is the indefinite article and does not have a longer form.
<'t> and <'k> are contractions of the words <het> `it' and <ek> `I'.
Having both contracted and uncontracted forms in the data could increase the perplexity of the language model.
This experiment looks at expanding the contractions <'t> and <'k> into their full forms <het> and <ek>.\footnote{The full and contracted forms may have differences in register or semantics, similar to English ``it is'' versus ``it's''.}

\subsection{Hausa \& Igbo Orthographic Standards}

Hausa and Igbo both have multiple orthographic standards, so we perform the same type of experiment on both languages.

Hausa is spoken in Niger and Nigeria, where both countries use a different orthographic standard \cite{coulmas1999blackwell}.
In Niger, the phoneme /\textipa{P}\textsuperscript{j}/ is written as <\includegraphics[width=0.09in]{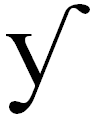}>, whereas in Nigeria the same phoneme is written as a sequence of <'> plus <y>, yielding <'y>.

Igbo officially uses the \textsubdot{O}nw\textsubdot{u} alphabet, developed in the early 1960s \cite{awde1999igbo}.
Due to apparent shortcomings in the \textsubdot{O}nw\textsubdot{u} alphabet for notating tone marking diacritics on typewriters and computers, the New Standard Alphabet was developed in 1976 \cite{oluikpe2014igbo}.
The New Standard Alphabet replaces \textsubdot{O}nw\textsubdot{u} <\textsubdot{o}> with <\"o>, <\textsubdot{u}> with <\"u>, and <\.{n}> with <\~n>.

Our experiments for these two languages involve normalizing each language to its two respective standards, either normalizing everything to the Niger/Nigeria standard for Hausa or to the \textsubdot{O}nw\textsubdot{u}/New Standard Alphabet standard for Igbo.
The end result will be a data set that uniformly follows one or the other standard, for each language.

\subsection{Somali \& Swahili Data Set Quality}

Not every language we consider has language-specific rewrite rules.
Somali and Swahili have relatively straightforward orthographies, and our research at this time does not suggest any obvious normalization experiments for these languages.
In this case, we are still interested in the quality of the data sets available for Somali and Swahili and so still run the text normalizer over them.

\section{Language-Specific Rule Results}

\begin{table*}[tbh!]
	\small
	\begin{center}
		\begin{tabu} to \linewidth {X[0.8,c]X[1.45,c]X[1,c]X[1,c]X[1,c]X[1,c]X[1,c]}\toprule
			\vspace{-1pt}\textbf{Language} & \vspace{-1pt}\textbf{Data Source} & \textbf{Base Perplexity} & \textbf{Experiment Perplexity} & \textbf{Raw Difference} & \textbf{Relative Difference} & \textbf{Difference from Median}\\\midrule

			%			Bambara	&	UD	&	646.75	&		&		&		&	\\\cmidrule(lr){1-7}

			%			\textbf{Wolof}	&	UD	&	2,905.63	&		&		&		&	\\\cmidrule(lr){1-7}

			%			\multirow{3}{*}{Yoruba}	&	UD	&	1,040.74	&		&		&		&	\\
			%			{}	&	LCC-wiki-10K	&	12,501.11	&		&		&		&	\\
			%			{}	&	OSCAR	&	602.03	&		&		&		&	\\\cmidrule(lr){1-7}
			%{}	&	OSCAR-dedup	&		&		&		&		&	\vspace{8pt}\\

			\multirow{3}{*}{Amharic}	&	\textbf{UD}	&	2,248.49	&	2,241.58	&	-6.91	&	-0.00302080	&	0.00298591\\
			{}	&	\textbf{LCC-wiki-30K}	&	58,005.51	&	57,112.44	&	-893.07	&	-0.00745546	&	0.00742057\\
			%			{}	&	LCC-wiki-10K	&	29,175.74	&	28,804.04	&	-371.69	&	-0.00957607	&	\\
			{}	&	\textbf{OSCAR}	&	52,936.50	&	51,901.28	&	-1,035.22	&	-0.01040426	&	0.01036937\\\cmidrule(lr){1-7}
			%{}	&	OSCAR-dedup	&		&		&		&		&	\vspace{8pt}\\

			\multirow{4}{*}{Zulu}	&	\textbf{LCC-mixed-30K}	&	85,290.26	&	85,270.72	&	-19.54	&	-0.00012419	&	0.00298591\\
			%			{}	&	LCC-mixed-10K	&	41,417.27	&	41,408.63	&	-8.64	&	-0.00016405	&	\\
			{}	&	LCC-news-30K	&	74,579.78	&	74,577.75	&	-2.04	&	-0.00001297	&	-0.00002192\\
			%			{}	&	LCC-news-10K	&	37,543.01	&	37,540.91	&	-2.10	&	-0.00003962	&	\\
			{}	&	\textbf{LCC-web-30K}	&	81,354.12	&	81,318.75	&	-35.37	&	-0.00024212	&	0.00020723\\
			%			{}	&	LCC-web-10K	&	39,281.59	&	39,274.27	&	-7.32	&	-0.00015008	&	\\
			{}	&	\textbf{LCC-web-za-30K}	&	86,582.27	&	86,558.96	&	-23.31	&	-0.00015269	&	0.00011780\\\cmidrule(lr){1-7}
			%			{}	&	LCC-web-za-10K	&	42,141.78	&	42,133.10	&	-8.68	&	-0.00017151	&	\\\cmidrule(lr){1-7}

			\multirow{3}{*}{Malagasy}	&	\textbf{LCC-web-30K	}&	28,188.62	&	28,170.46	&	-18.16	&	-0.00010813	&	0.00007324\\
			%			{}	&	LCC-web-10K	&	15,237.62	&	15,231.35	&	-6.27	&	-0.00011040	&	\\
			{}	&	\textit{LCC-wiki-30K}	&	16,859.84	&	16,859.84	&	0	&	0	&	-0.00003489\\
			%			{}	&	\textit{LCC-wiki-10K}	&	7,081.93	&	7,081.93	&	0	&	0	&	\\
			{}	&	OSCAR	&	12,438.60	&	12,436.92	&	-1.68	&	-0.00001568	&	-0.00001921\\\cmidrule(lr){1-7}
			%{}	&	OSCAR-dedup	&		&		&		&		&	\vspace{8pt}\\

			\multirow{10}{*}{Afrikaans}	&	\textit{UD}	&	3,457.22	&	3,457.22	&	0	&	0	&	-0.00003489\\
			{}	&	LCC-mixed-30K	&	31,829.22	&	31,824.21	&	-5.01	&	-0.00002628	&	-0.00000861\\
			{}	&	\textit{LCC-mixed-10K}	&	17,050.93	&	17,050.93	&	0	&	0	&	-0.00003489\\
			{}	&	LCC-news-30K	&	26,366.78	&	26,366.17	&	-0.62	&	-0.00000285	&	-0.00003204\\
			%			{}	&	LCC-news-10K	&	14,958.55	&	14,956.31	&	-2.24	&	-0.00003017	&	\\
			{}	&	LCC-newscrawl-30K	&	30,392.04	&	30,391.76	&	-0.27	&	-0.00000133	&	-0.00003356\\
			%			{}	&	LCC-newscrawl-10K	&	16,529.73	&	16,529.38	&	-0.35	&	-0.00000498	&	\\
			{}	&	LCC-web-30K	&	29,840.21	&	29,839.93	&	-0.28	&	-0.00000143	&	-0.00003346\\
			{}	&	\textit{LCC-web-10K}	&	15,645.45	&	15,645.45	&	0	&	0	&	-0.00003489\\
			{}	&	LCC-wiki-30K	&	38,254.72	&	38,254.42	&	-0.29	&	-0.00000139	&	-0.00003350\\
			{}	&	\textit{LCC-wiki-10K}	&	20,247.26	&	20,247.26	&	0	&	0	&	-0.00003489\\
			{}	&	OSCAR	&	16,656.40	&	16,656.20	&	-0.21	&	-0.00000120	&	-0.00003369\\\cmidrule(lr){1-7}
			%		{}	&	OSCAR-dedup	&		&		&		&		&	\vspace{8pt}\\

			Hausa	&	\textit{AC}	&	5298.80	&	5298.80	&	0	&	0	&	-0.00003489\\\cmidrule(lr){1-7}

			Igbo	&	\textit{AC}	&	5928.83	&	5928.83	&	0	&	0	&	-0.00003489\\\bottomrule
			
		\end{tabu}
		\caption{Results of normalization experiments on perplexity. \textbf{Bold} values for Data Source indicate conditions where the relative decrease in perplexity was greater than the median decrease in perplexity. \textit{Italic} values for Data Source indicate conditions where there was no decrease in perplexity.}
		\label{table:exp-results}
	\end{center}
\end{table*}

Table~\ref{table:exp-results} shows the results of our language-specific rule experiments on the perplexity of the language model.
The patterns observed for the language-specific rules are the following:

\paragraph{Amharic} Reducing grapheme overdifferentiation reduces perplexity for all data sets tested.

\vspace{-6pt}
\paragraph{Zulu} Removing hyphens between noun classifiers and vowel-initial loanwords reduces perplexity for all data sets tested.

\vspace{-6pt}
\paragraph{Malagasy} No change in perplexity for LCC-wiki corpus (no occurrences of standalone <@> and only three of <\~n> in rejected sentences).
Decrease in perplexity in LCC-web and OSCAR corpora.

This variability could be due to the informal nature of the phenomena we looked at.
Informal abbreviations like <@> may not show up on Wikipedia, but they may be more likely in the miscellaneous sources found in the LCC-web and OSCAR corpora.

\vspace{-6pt}
\paragraph{Afrikaans} No change in perplexity for the UD, LCC-mixed-10K, LCC-web-10K, and LCC-wiki-10K corpora.
Once we increase to 30K for those LCC corpora, we do see a small decrease in perplexity.
We observe a decrease in perplexity for the OSCAR, LCC-news, and LCC-newscrawl corpora, even for the 10K corpora.

Similar to Malagasy, some of these differences could be due to the nature of the Afrikaans contractions and the domains of the data sets.
The Afrikaans UD data set is composed of legal documents.
If the contractions are informal, it should be no surprise that we don't encounter them in the UD data.
We find contractions in the 10K news and newscrawl corpora, but not in the 10K web corpus.
If news is written in more formal language and web sources are less formal, we would expect the opposite pattern.

\begin{table*}[tbh!]
	\small
	\begin{center}
		\begin{tabu} to \linewidth {ccccc}\toprule
			\textbf{Language} & \textbf{Data Source} & \textbf{Sentences Kept} & \textbf{Sentences Rejected} & \textbf{Percent Rejected}\\\midrule

			%			Bambara	&	UD	&	1,026	&	30	&	2.92\\\cmidrule(lr){1-5}

			%			\textbf{Wolof}	&	UD	&	1,188	&	56	&	4.71\\\cmidrule(lr){1-5}

			%			\multirow{3}{*}{Yoruba}	&	UD	&	294	&	24	&	7.55\\
			%			{}	&	LCC-wiki-10K	&	6,112	&	3,888	&	38.88\\
			%			{}	&	OSCAR	&	487	&	201	&	29.22\\\cmidrule(lr){1-5}
			%			{}	&	OSCAR-dedup	&		&		&	\\\cmidrule(lr){1-5}

			\multirow{3}{*}{Amharic}	&	UD	&	1,047	&	27	&	2.51\\
			{}	&	LCC-wiki-30K	&	21,165	&	8,835	&	29.45\\
			%			{}	&	LCC-wiki-10K	&	7,095	&	2,905	&	29.05\\
			{}	&	OSCAR*	&	15,424	&	14,432	&	48.34\\\cmidrule(lr){1-5}
			%			{}	&	OSCAR-dedup	&		&		&	\\\cmidrule(lr){1-5}

			\multirow{4}{*}{Zulu}	&	LCC-mixed-30K	&	26,547	&	3,453	&	11.51\\
			%			{}	&	LCC-mixed-10K	&	8,865	&	1,135	&	11.35\\
			{}	&	LCC-news-30K	&	24,659	&	5,341	&	17.80\\
			%			{}	&	LCC-news-10K	&	8,200	&	1,800	&	18.00\\
			{}	&	LCC-web-30K	&	25,943	&	4,057	&	13.52\\
			%			{}	&	LCC-web-10K	&	8,640	&	1,360	&	13.60\\
			{}	&	LCC-web-za-30K	&	26,260	&	3,740	&	12.47\\\cmidrule(lr){1-5}
			%			{}	&	LCC-web-za-10K	&	8,795	&	1,205	&	12.05\\\cmidrule(lr){1-5}

			\multirow{3}{*}{Malagasy}	&	LCC-web-30K	&	24,442	&	5,558	&	18.53\\
			%			{}	&	LCC-web-10K	&	8,158	&	1,842	&	18.42\\
			{}	&	LCC-wiki-30K	&	19,252	&	10,748	&	35.83\\
			%			{}	&	LCC-wiki-10K	&	6,357	&	3,643	&	36.43\\
			{}	&	OSCAR*	&	10,942	&	1,591	&	12.69\\\cmidrule(lr){1-5}
			%			{}	&	OSCAR-dedup	&		&		&	\\\cmidrule(lr){1-5}

			\multirow{7}{*}{Afrikaans}	&	UD	&	1,249	&	66	&	5.02\\
			{}	&	LCC-mixed-30K	&	25,509	&	4,491	&	14.97\\
			%			{}	&	LCC-mixed-10K	&	8,529	&	1,471	&	14.71\\
			{}	&	LCC-news-30K	&	27,144	&	2,856	&	9.52\\
			%			{}	&	LCC-news-10K	&	9,091	&	909	&	9.09\\
			{}	&	LCC-newscrawl-30K	&	27,071	&	2,929	&	9.76\\
			%			{}	&	LCC-newscrawl-10K	&	9,025	&	975	&	9.75\\
			{}	&	LCC-web-30K	&	26,069	&	3,931	&	13.10\\
			%			{}	&	LCC-web-10K	&	8,701	&	1,299	&	12.99\\
			{}	&	LCC-wiki-30K	&	26,308	&	3,692	&	12.31\\
			%			{}	&	LCC-wiki-10K	&	8,787	&	1,213	&	12.13\\
			{}	&	OSCAR*	&	22,125	&	3,708	&	14.35\\\cmidrule(lr){1-5}
			%			{}	&	OSCAR-dedup	&		&		&	\\\cmidrule(lr){1-5}

			Hausa	&	AC	&	41,276	&	8,724	&	17.45\\\cmidrule(lr){1-5}

			Igbo	&	AC	&	40,762	&	9,238	&	18.48\\\cmidrule(lr){1-5}

			\multirow{3}{*}{Somali}	&	LCC-newscrawl-30K	&	26,563	&	3,437	&	11.46\\
			%			{}	&	LCC-newscrawl-10K	&	8,837	&	1,163	&	11.63\\
			{}	&	LCC-wiki-10K	&	8,378	&	1,613	&	16.13\\
			{}	&	OSCAR	&	8	&	171	&	95.53\\\cmidrule(lr){1-5}
			%			{}	&	OSCAR-dedup	&		&		&	\\\cmidrule(lr){1-5}

			\multirow{3}{*}{Swahili}	&	LCC-newscrawl-swa-10K	&	8,796	&	1,204	&	12.04\\
			{}	&	LCC-wiki-swa-30K	&	25,743	&	4,257	&	14.19\\
			%			{}	&	LCC-wiki-swa-10K	&	8,582	&	1,418	&	14.18\\
			{}	&	LCC-wiki-swh-30K	&	25,457	&	4,543	&	15.14\\\bottomrule
			%			{}	&	LCC-wiki-swh-10K	&	8,511	&	1,489	&	14.89\\\bottomrule
			
		\end{tabu}
		\caption{Kept and rejected sentences from each data source for each language.
			For Amharic, Malagasy, and Afrikaans, only a subset of the OSCAR data was used due to computational constraints.
			The OSCAR data was limited to 10,000 lines, with each line consisting of one or more sentences.}%, thus the variable number of total sentences for each of these languages.}
		\label{table:results-reject}
	\end{center}
\end{table*}

\vspace{-6pt}
\paragraph{Hausa \& Igbo} We observe no difference in perplexity between the base and experiment conditions.
Looking into the data, we find that the Hausa corpus has no occurences of the grapheme <\includegraphics[width=0.09in]{y-hook.png}>.
Relatedly, the Igbo corpus has no occurrences of New Standard Alphabet graphemes <\"o>, <\"u>, or <\~n>, so normalizing from the New Standard Alphabet has no effect.

Lastly, we observe each experiment's relative perplexity difference from the median.
This value can tell us which experiments are doing more work compared to others.
We find that our Amharic, Zulu, and Malagasy experiments tend to have a greater relative decrease in perplexity than the median, and Afrikaans, Hausa, and Igbo have less of a relative decrease in perplexity than the median.

\section{Data Set Quality Results}

Table~\ref{table:results-reject} shows the results of our experiments on data set quality for each experimental condition.
Table~\ref{table:results-reject-summary} shows a summary of these results grouped by data source.

\begin{table}[tbh!]
	\begin{center}
		\begin{tabu} to \linewidth {lrr}\toprule
			\multicolumn{1}{c}{\textbf{Data Source}} & \multicolumn{1}{c}{\textbf{Average}} & \multicolumn{1}{c}{\textbf{Median}}\\\midrule
			UD	&	4.71	&	4.54\\
			LCC-all	&	13.60	&	16.51\\
			LCC-mixed	&	13.14	&	13.11\\
			LCC-news	&	13.60	&	13.66\\
			LCC-newscrawl	&	10.93	&	11.46\\
			LCC-web	&	14.34	&	13.31\\
			LCC-wiki	&	22.38	&	15.64\\
			OSCAR	&	29.22	&	40.03\\
			AC	&	17.96	&	17.96\\\bottomrule
		\end{tabu}
		\caption{Average and median percentage of rejected sentences from each data source.}
		\label{table:results-reject-summary}
	\end{center}
\end{table}

The patterns observed for each data source are the following.
\textbf{UD} shows a very low proportion of sentences rejected.
This is likely because UD is very clean data. %, but the domains of the sources vary as well.
\textbf{LCC} shows a higher proportion of sentences rejected, varying by genre.
The LCC newscrawl data has the lowest proportion of rejections out of the LCC genres, and the LCC wiki data has the highest proportion.
\textbf{OSCAR} has a higher, variable proportion of sentences rejected.
The Somali OSCAR corpus had highest proportion of rejected sentences out of any data sets we looked at (95.53\%).
\textbf{AC} was only used for Hausa \& Igbo, so statistics might not be very informative.
However, the AC data seems to have a higher proportion of rejections than all of the other data sources except for OSCAR and LCC-wiki.

We also observe variability within each data source.
For example, we saw that the LCC wiki data had the most rejections of the LCC genres.
When we look at each language, however, we see that the Afrikaans, Somali, and Swahili LCC wiki results are in the low to mid teens.
The Yoruba, Amharic, and Malagasy LCC wiki results---all above 29\% rejected---bring the average and median for LCC wiki up.
Similarly, the Malagasy and Afrikaans OSCAR results are also in the low teens, with Yoruba and Somali bringing up the overall OSCAR average and median.

%\vspace{-6pt}
\section{Future Work \& Conclusion}

%\vspace{-6pt}
Based on our results, several areas for future work can be identified.

First, we observe that some data sets did not show any change in perplexity between the base and experiment condition, while other data sets for the same language do (e.g. Afrikaans).
Looking deeper, we see that the genre or style of the data and the type of change we are making play an important role.
For Afrikaans, the contractions <'t> and <'k> are informal.
Thus, we should not expect to find many occurrences of them in a corpus of legal documents, which is exactly what we see with the Afrikaans UD data.
Taking the genre/domain of the data into account can help inform whether language-specific changes will be effective in normalizing the data.

Second, we saw no change in perplexity for the Hausa or Igbo experiments.
Again, this comes down to the data sources.
For both of these languages, we only used the An Cr\'ubad\'an data.
However, both languages do have a Wikipedia, so future work could use that to potentially create a more representative data set.
However, one goal of this project was to investigate \textit{existing} data sets, which is why we did not use Wikipedia to make our own Hausa or Igbo corpora in the first place.

Third, while we knew that using perplexity to evaluate the language model here was precarious, our findings suggest a new use case.
Even if we are unsure if the change in perplexity is actually better or worse, we can still use it to determine if our changes have applied or not.
Recall the Hausa and Igbo cases, where our language-specific rules---no matter how well-intentioned they were---did nothing.
This metric can be used as a first step in evaluating the usefulness of a data set for a downstream task.
If the goal is to replicate how speakers of a language actually type, then perhaps using legal documents or religious texts as training data will not yield the desired results.

In conclusion, this paper reports on three main topics.
First, text normalization is important, especially for low-resource languages with millions of speakers.
Second, linguistic analysis is an important part of text normalization.
Without it, we could not implement the language-specific normalization rules described above.
This is particularly important for African languages, which have complexities not commonly found in the more-studied languages of e.g.\ Europe, such as multiple standard orthographies.
Third, existing open source data sets can help inform design decisions for low-resource language technology.

%\section*{Acknowledgements}
%Acknowledgements go here!

\bibliographystyle{acl_natbib}
\bibliography{textnorm}

\end{document}